\documentclass{article}

\usepackage{arxiv}
\usepackage{amsmath,amssymb,amsthm}
\usepackage[utf8]{inputenc} 
\usepackage[T1]{fontenc}    
\usepackage{url}            
\usepackage{booktabs}       
\usepackage{amsfonts}       
\usepackage{nicefrac}       
\usepackage{microtype}      
\usepackage{atbegshi}%
\usepackage[utf8]{inputenc}
\usepackage{subcaption}
\usepackage{csquotes}
\usepackage{xcolor}
\usepackage{graphicx}
\usepackage{booktabs}
\usepackage{algorithm}
\usepackage[noend]{algorithmic}
\usepackage{float}
\usepackage{soul}
\usepackage{natbib}

\usepackage{tikz}
\usepackage{pgfplots}
\pgfplotsset{compat=1.17}

\sethlcolor{lightgray}

\DeclareMathOperator*{\argmax}{arg\,max}

\title{Evolutionary Diversity Optimisation for The Traveling Thief Problem}

\author{
 Adel Nikfarjam \\
Optimisation and Logistics\\School of Computer Science\\The University of Adelaide\\
  \texttt{adel.nikfarjam@adelaide.edu.au} \\
  \And
 Aneta Neumann \\
Optimisation and Logistics\\School of Computer Science\\The University of Adelaide\\
  \texttt{aneta.neumann@adelaide.edu.au} \\
    \And
 Frank Neumann \\
Optimisation and Logistics\\School of Computer Science\\The University of Adelaide\\
  \texttt{frank.neumann@adelaide.edu.au} \\}
\begin{document}
\maketitle
\begin{abstract}
There has been a growing interest in the evolutionary computation community to compute a diverse set of high-quality solutions for a given optimisation problem. This can provide the practitioners with invaluable information about the solution space and robustness against imperfect modelling and minor problems' changes. It also enables the decision-makers to involve their interests and choose between various solutions. In this study, we investigate for the first time a prominent multi-component optimisation problem, namely the Traveling Thief Problem (TTP), in the context of evolutionary diversity optimisation. We introduce a bi-level evolutionary algorithm to maximise the structural diversity of the set of solutions.
Moreover, we examine the inter-dependency among the components of the problem in terms of structural diversity and empirically determine the best method to obtain diversity. We also conduct a comprehensive experimental investigation to examine the introduced algorithm and compare the results to another recently introduced framework based on the use of Quality Diversity (QD). Our experimental results show a significant improvement of the QD approach in terms of structural diversity for most TTP benchmark instances.  
\end{abstract}

%
%


\section{Introduction}
Evolutionary Algorithms (EAs) are traditionally used to find a high quality solution, ideally optimum/near-optimum for a given optimisation problem. The diversification of a set of high-quality solutions has gained increasing attention in the literature of evolutionary computation in recent years. These studies are dominated mainly by multi-modal optimisation. They aim to explore the fitness landscape to find niches, usually through a diversity preservation mechanism. Moreover, several studies can be found focusing on exploring niches in a feature space. The paradigm is called Quality Diversity (QD), where the objective is to find a set of high-quality solutions that differ in terms of some user-defined features. QD has been mainly applied to the areas of robotics \cite{RakicevicCK21, ZardiniZZIF21, Cully2201}, and games \cite{SteckelS21, FontaineTNH20, FontaineLKMTHN21}.

Evolutionary Diversity Optimisation (EDO) is another concept in this area. In contrast to the previous paradigms, EDO explicitly seeks to maximise the structural diversity of solutions, generally subject to a quality constraint. \citet{ulrich2011maximizing} first defined the outline of EDO. Afterwards, the concept has been utilised to evolve a diverse set of images and the Traveling Salesperson Problem's (TSP) instances in \cite{alexander2017evolution,doi:10.1162/evcoa00274}. For the same purposes, the star discrepancy and indicators from the multi-objective optimisation frameworks have been studied in \cite{neumann2018discrepancy} and \cite{neumann2019evolutionary}, respectively. The use of distance-based diversity measures and entropy have been studied in \cite{viet2020evolving, NikfarjamBN021a} to generate a diverse set of solutions for the TSP. \citet{NikfarjamB0N21b} introduced a modified Edge Assembly Crossover (EAX) to achieve higher diversity in TSP tours. Moreover, EDO has been studied in the context of knapsack problem~\cite{BossekN021KP}, minimum spanning tree problem~\cite{Bossek021tree}, quadratic assignment problem~\cite{DoGN021} and the optimisation of monotone sub-modular functions~\cite{NeumannB021}.  

Real-world optimisation problems often include several sub-problems interacting, where each sub-problem impacts not only the quality but also the feasibility of solutions of others. These kinds of problems are called multi-component optimisation problems. Traveling Thief Problem (TTP) can be classified into this category. TTP is the integration of TSP and the Knapsack Problem (KP), where the traveling cost between two cities depends on the distance between the cities and the weight of items collected \cite{BonyadiMB13}. A wide range of solution approaches have been proposed to TTP that includes co-evolutionary strategies \cite{BonyadiMPW14, YafraniA15}, swarm intelligence approaches \cite{Wagner16, ZouariAT19}, simulated annealing~\cite{YafraniA18}, and local search heuristics \cite{PolyakovskiyB0MN14, MaityD20}. More recently, \citet{NikfarjamMap} introduced a Map-elite based algorithm to compute a set of high-quality solutions exploring niches in a feature space. They showed that the algorithm is capable of improving the best-known solutions for several benchmark instances.    

\subsection{Our Contribution}
In this study, we investigate the EDO in the context of TTP. Several advantages can be found for having a diverse set of high-quality solutions for TTP. First, we can study the inter-dependency of the sub-problems in terms of structural diversity and find the best method to maximise it. Second, EDO provides us with invaluable insight into the solutions space. For example, it can show which elements of an optimal/near-optimal solution can be replaced easily and which elements are irreplaceable. Finally, it brings about robustness against the minor changes. 

To the best of our knowledge, this study is the first to investigate EDO in the context of a multi-component problem. We first establish a method to calculate the structural diversity of TTP solutions. Then, we introduced a bi-level EA to maximise the diversity. The first level involves generating the TSP part of a TTP solution, whereby the second level is an inner algorithm to optimise the KP part of the solution with respect to the first part. Then, an EDO-based survival selection is exercised to maximise the diversity. We first examine the impact that incorporating different inner algorithms into the EA can make on the diversity of the solutions.
Moreover, We empirically study the inter-dependency between the sub-problems and show how focusing on the diversity of one sub-problem affects the other's and determine the best method to obtain diversity. Interestingly, the results indicate that focusing on overall diversity brings about greater KP diversity than solely emphasising KP diversity. In addition, we compare the set of solutions obtained from the introduced algorithm with a recently developed QD-based EA in terms of structural diversity. The results show that the introduced bi-level EA can bring higher structural diversity for most test instances. We also conduct a simulation test to evaluate the robustness of populations obtained from the two algorithms against changes in the problem.

The remainder of the paper is structured as follows. We formally define the TTP problem and the diversity for a set of TTP solutions in Section \ref{Sec:prob_def}. In Section \ref{Sec:alg}, We introduce the two-stage EA. A comprehensive experimental investigation is conducted in Section \ref{Sec:exp}. Finally, we finish with some concluding remarks. 
\section{Problem Definition}
\label{Sec:prob_def}
TTP is defined on the aggregation of the TSP and the KP. The TSP is formed by a complete graph $G =(V, E)$, where $V$ is a set of cities, and $E$ is pairwise edges that connect the cities. We denote the size of the cities set by $|V| = n$. There is also a non-negative distance $d(e)$ associated with each edge $e = (u, v) \in E$. In TSP, the goal is to find a permutation (a tour) $x: V \to V$ that minimise a cost function. The KP is defined on a set of items $I$, where $|I| = m$. Each item $i$ associated with a weight $w_i$ and a profit $p_i$. In KP, the objective is to find a selection of items $y = (y_1, \cdots, y_m)$ that maximise profit complied with the wight of the selected items not exceeding a capacity of $W$. Note that $y_i$ is a binary variable equal to 1 if i is included in packing list $y$; otherwise is equal to 0.

The TTP is defined on the graph $G$ and the set of items $I$. However, the items are scattered over the cities. Each city $j \in V \setminus v_1$ has a set of item $M_j$, where $M_j \subset I$. the thief visits each city exactly once and collect some items into the knapsack. Moreover, a rent of $R$ should be paid for the knapsack per time unit, and $\nu_{\max}$ and $\nu_{\min}$ are the maximum and minimum speeds that the thief can travel, respectively. In the TTP, we aim to compute a solution including a tour $x$ and a packing list $y$ maximising the following objective function $z(x, y)$:
\begin{align*}
& z(x,y) = \sum_{j=1}^{m} p_j y_j - R \left( \frac{d(x_n, x_1)}{\nu_{max}-\nu W_{x_n}} + \sum_{i=1}^{n-1} \frac{d(x_i, x_{i+1})}{\nu_{max}-\nu W_{x_i}} \right)\\
&\text{subject to } \sum_{j=1}^{m} w_j y_j \leq W \quad y_j\in\{0,1\}.
\end{align*}
where $W_{x_i}$ is the cumulative weight of the items collected from the start of the tour up to city $x_i$, and $\nu = \frac{\nu_{max}-\nu_{min}}{W}$ is a constant.

This study aims to compute a diverse set of TTP solutions that all comply with a minimum quality threshold but differ in terms of structural properties. In other words, the objective is to maximise the diversity of the set of solutions subject to a quality constraint. Let denote the set of TTP solutions by $P = \{p_1, \cdots, p_\mu\}$, where $|P| = \mu$. Therefore, we can formally formulate the problem as:

\begin{align*}
& Max H(P) \\
&\text{subject to }\\ 
& z_p \geq (1 - \alpha)z^* & \forall p \in P\\
&\sum_{j=1}^{m} w_{j} y_{jp} \leq W & \forall p \in P \quad y_{jp}\in\{0,1\}
\end{align*}
Where $H(P)$ is a measure quantifying the diversity of $P$, $z^*$ is the optimal or the best-known value of $z$ for a given TTP instance, $\alpha$ is the acceptable quality threshold, and $y_{jp}$ shows $y_j \in p$. In line with most of the studies in EDO literature, we assumed that the optimal or a high quality solution of TTP instances are already known.   

\subsection{Diversity in TTP}
\label{Sec:div}
To maximise the diversity, we require a measure to quantify the diversity of a set of solutions. As mentioned, a TTP solution includes two different parts, a tour and a packing list. That means a function is required to calculate the structural diversity of tours and another one for packing lists. We adopt the well-known information-theoretic concept of entropy for this purpose.

We employ the diversity measure based on entropy from \cite{NikfarjamBN021a, NikfarjamB0N21b} to compute the entropy of the tours. Let $P$ be a set of TTP solutions. Here, the diversity is defined on the proportion of edges including in $E(P)$, where $E(P)$ is the set of edges included in $P$. The edge entropy of $P$ can be calculated from:
\begin{align*}
H_{e}(P) = \sum_{e \in E(P)} h(e) \text{ with } h(e)=-\left(\frac{f(e)}{2n\mu}\right)\ln{\left(\frac{f(e)}{2n\mu}\right)}.
\end{align*}
where $h(e)$ is the contribution of an edge $e \in E$ to the entropy, and $f(e)$ is the number of tours in $P$ including $e$. The contribution of edges with zero frequency is equal to zero ($h(e) = 0 \iff f(e) = 0$). $2n\mu$ is the summation of the frequency of all edges over the population.   

The same concept is adopted for calculation of the entropy of items. The diversity of packing list on the proportion of items being included in $P(I)$, where $P(I)$ is the set of items included in $P$. The item entropy of $P$ can be compute from:
\begin{align*}
H_i(P) = \sum_{i \in P(I)} h(i) \text{ with } h(i)=-\left(\frac{f(i)}{\sum_{i \in I} f(i)}\right)\ln{\left(\frac{f(i)}{\sum_{i \in I} f(i)}\right)}.
\end{align*}
where $h(i)$ is the contribution of an item $i \in I$ to the entropy, and $f(i)$ is the number of packing lists in $P$ including $i$. The contribution of items with zero frequency is equal to zero ($h(i) = 0 \iff f(i) = 0$).

A simple way to calculate the overall entropy is to sum up the entropy of edges and items. This is because $H_e$ and $H_i$ are basically the summation of contribution of edges and items. Therefor, we have: $H(P) = H_e(P) + H_i(P)$

\section{Bi-level Evolutionary Algorithm}
\label{Sec:alg}
We introduce a bi-level EA to compute a diverse set of TTP solutions. The EA is started with an initial population that all individuals complying the quality constraint. The procedure to construct such a population will be explained later. Having selected two tours uniformly at random, the EA generates a new tour by crossover. Then, an inner algorithm is initiated to compute a corresponding packing list for the new tour in order to have a complete TTP solution; we refer the inner algorithms as the KP operators. If the TTP score of the new solution is higher than minimum requirement, it will be added to the population; otherwise, it will be discarded. Finally, an individual with minimum contribution to the diversity of population will be discarded if the size of population is $\mu + 1$. These steps are continued until a termination criterion is met. Algorithm \ref{alg:diversity_maximizing_EA} outlines the bi-level EA. In this study, we employ EAX as crossover and Dynamic Programming~(DP) or alternatively $(1+1)$EA as the KP operators. These operators are shown in \cite{NikfarjamMap} capable of computing high-quality TTP solutions efficiently.
\subsection{The Edges Assembly Crossover (EAX)}
\begin{figure}[t]
    \centering
    \includegraphics[width=\columnwidth]{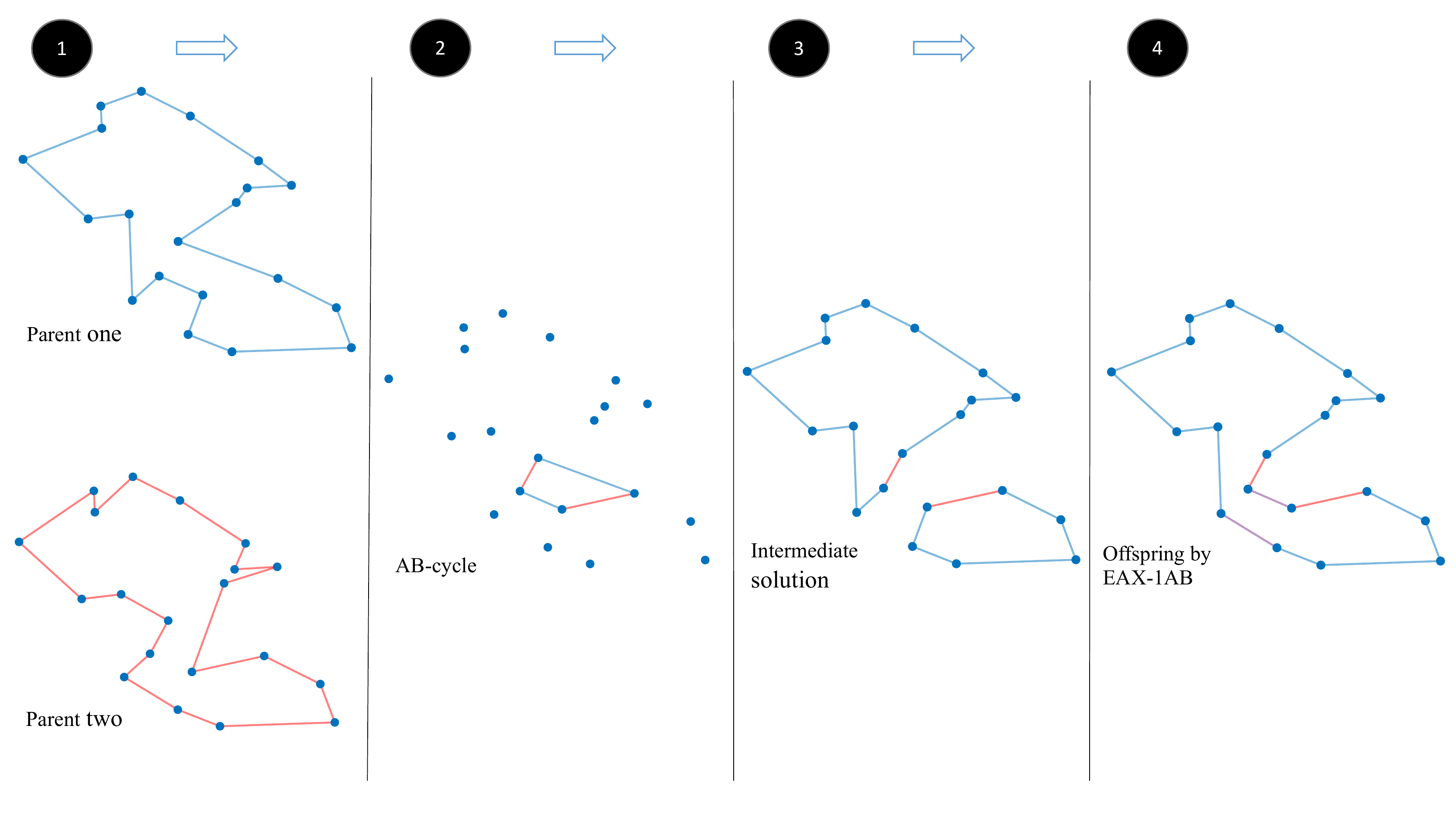}
    \caption{The representation of the steps to implement EAX \cite{NikfarjamMap}}
    \label{fig:example_eax_1ab}
\end{figure}

The EAX is known to yield decent TSP tours and the GA using the EAX \cite{nagata2013powerful} is a high-performing EA in solving TSP. Several variations of the EAX can be found in the literature. This study utilises the EAX-1AB for its simplicity and efficiency compared to the other variants. Since we only use EAX-1Ab, we refer it as the EAX. As shown in Figure \ref{fig:example_eax_1ab}, the crossover is formed by three steps as follows:
\begin{itemize}
    \item AB-cycle: forming an AB-cycle from two tours by alternatively selecting edges from first and second tours until a cycle is formed (Fig \ref{fig:example_eax_1ab}.2). 
    \item Intermediate Solution: Copying all edges from the first tour; then removing the Ab-cycle's edges belonging to the first tour, and adding the other edges of the AB-cycle. (Fig \ref{fig:example_eax_1ab}.3).
    \item Completing the Tour: Connecting sub-tours of the intermediate solution to have one complete tour (Fig \ref{fig:example_eax_1ab}.4).
\end{itemize}
We require a $4$-tuples of edges to connect two sub-tours, one edge from each sub-tour to be removed and two new edges to connect each end of the discarded edges. We first select the sub-tour $r$ with the minimum edge number. afterwards, we determine the $4$-tuples of edges such that $\{e_1, e_2, e_3, e_4\} = \arg \min \{-d(e_1)-d(e_2)+d(e_3)+d(e_4)\}$. Note that  $e_1 \in E(r)$ and  $e_2 \in E(t) \setminus E(r)$, where $E(r)$ and $E(t)$ denote the set of edges included in sub-tour $r$ and the intermediate solution $t$, respectively. The interested readers are referred to \cite{nagata2013powerful} for more details about implementation of the EAX.
\subsection{The KP operators}
Having generated a new tour, we employ a DP approach to compute a packing list for the generated tour and form a high-quality TTP solution. DP is classically utilised to solve the KP. \citet{PolyakovskiyB0MN14} introduced a DP approach to solve the Packing While Traveling problem (PWT), which is a simplified variant of the TTP. the difference between the PWD and the TTP is that the tour is fixed in the PWD and a solution only includes a packing list. Here, we use the same DP to compute the optimal packing list for the tour generated by EAX.      

The DP includes a table $\beta$ size of $(W\times m)$. Here, we are processing the items based on the order of the appearance of their corresponding nodes in the tour. In other words, $I_i$ is processed sooner than $I_j$ if $I_i$ belongs to a node visited prior $I_j$. If $I_i$ and $I_j$ belong to the same node, they are processed based on the their indices. The $\beta_{i,j}$ shows the maximal profit among all combinations of items $I_k$ with $I_k \preceq I_i$ result in the weight exactly equal to $j$. If there are no combinations bringing about the weight $j$, $\beta_{i,j}$ is set to $-\infty$.    

Let denote the profit of an empty set and the set including only item $I_i$ by $B(\emptyset)$ and $B(I_i)$, respectively. For the first row of the table we have:
$$\beta_{i,0} = B(\emptyset),\quad\beta_{i,w_i} = B(I_i),\quad\beta_{i,j} = -\infty, \forall j\notin \{0,w_i\}$$
Let $I_k$ be the predecessor of $I_i$. We can calculate $\beta_{i,j}$ from $\max(\beta_{k,j}, T)$, where $T$ is computed as follows:
$$T = \beta_{k,j-w_i}+p_i-R\sum_{l=1}^n d_l\left(\frac{1}{\nu_{max}-\nu j}-\frac{1}{\nu_{max}-\nu j-w_i}\right)$$
The $\max_j \beta_{m,j}$ is the maximum profit that we can get from the given tour. As mentioned, the DP results in the optimal packing list for the given tour; However, the run-time of the DP is quit long. Moreover, The DP is an exact solver since we aim to increase the diversity of items, it would be interesting to compare the results of the DP with a random algorithm such as an EA. Thus, we introduce a simple $(1+1)$EA as an alternative for the DP. 

The $(1+1)$EA is initialised with the packing list of the first parent used for generating the tour. Then, a new packing list is generated by mutation. If the new packing list results in a higher profit $z(x, y)$, it will be replaced with the old one; otherwise, it will be discarded. These steps are continued until a termination criterion is met. Here, we consider the bit-flip mutation. Each bit is independently filliped by the probability of $(1/m)$.

\begin{algorithm}[t!]
\begin{algorithmic}[1]
\REQUIRE{Population $P$, minimal quality threshold $c_{min}$}
\WHILE{termination criterion is not met}
\STATE Choose $x_1$ and $x_2 \in P$ uniformly at random, and generate one tour $x_3$ by crossover.\\ 
\STATE Generate a corresponding packing list ($y_3$) by a KP operator to have a complete TTP solution ($p_3(x_3, y_3)$) 
\IF{$z(p_3) \geq z_{min}$}
\STATE Add $p_3$ to $P$.
\ENDIF
\IF{$|P| = \mu+1$}
\STATE Remove one individual $p$ from $P$, where $p = \argmax_{q\in P} H(P \setminus \{q\})$.
\ENDIF
\ENDWHILE
\end{algorithmic}
\caption{Two-stage-EA}
\label{alg:diversity_maximizing_EA}
\end{algorithm}
\subsection{Initial Population}
As mentioned, we assumed that we know the optimal/near-optimal solution for given TTP instances; such an assumption is in line with most studies in the literature of EDO. The procedure is initialised with a single high-quality solution solution $p$ in $P$, where $z(p) \in ((1 - \alpha)z^* z^*)$. First, an individual $p \in P$ is selected uniformly at random. Then, the tour of the individual ($p(x)$) is mutated by 2-OPT, which is a well-known random neighborhood search in TSP. Afterwards, we compute a packing list $y'$ by KP and match it with the mutated tour $x'$ to have a TTP solution $p'$. If $p'$ complies the quality constraint, it will be added to $P$; otherwise, it is discarded. We continue these steps until $|P| = \mu$. Note that We used the algorithm introduced in \cite{NikfarjamMap} to obtain the initial $p$. Algorithm \ref{alg:Initial} outlines the initialising procedure.  

\begin{algorithm}[t!]
\begin{algorithmic}[1]
\REQUIRE{A TTP solution $p$ complying the quality criterion, population size $\mu$}
\WHILE{$|P|<\mu$}
\STATE Choose $p \in P$ uniformly at random, generate a tour $x'$ by mutation.\\ 
\STATE Compute a packing list $y'$ by the DP to form $p'(x',y')$) 
\IF{$z(p') \geq z_{min}$}
\STATE Add $p'$ to $P$.
\ENDIF
\ENDWHILE
\end{algorithmic}
\caption{Initial Population Procedure}
\label{alg:Initial}
\end{algorithm}
\begin{table}[t]
\renewcommand{\tabcolsep}{4pt}
    \centering
    \begin{tabular}{c l }
    \toprule
    Number & Original Name \\
    \midrule
         01 & eil51\_n50\_bounded-strongly-corr\_01\\
         02 & eil51\_n150\_bounded-strongly-corr\_01 \\
         03 & eil51\_n250\_bounded-strongly-corr\_01 \\
         04 & eil51\_n50\_uncorr-similar-weights\_01\\
         05 & eil51\_n150\_uncorr-similar-weights\_01 \\
         06 & eil51\_n250\_uncorr-similar-weights\_01\\
         07 & eil51\_n50\_uncorr\_01 \\
         08 & eil51\_n150\_uncorr\_01 \\ 
         09 & eil51\_n250\_uncorr\_01 \\
         10 & pr152\_n151\_bounded-strongly-corr\_01 \\
         11 & pr152\_n453\_bounded-strongly-corr\_01 \\
         12 & pr152\_n151\_uncorr-similar-weights\_01 \\
         13 & pr152\_n453\_uncorr-similar-weights\_01 \\
         14 & pr152\_n151\_uncorr\_01 \\
         15 & pr152\_n453\_uncorr\_01 \\
         16 & a280\_n279\_bounded-strongly-corr\_01 \\
         17 & a280\_n279\_uncorr-similar-weights\_01 \\
         18 & a280\_n279\_uncorr\_01\\
         \bottomrule
    \end{tabular}
    \caption{The names of the TTP instances are used in the paper.}
    \label{tab:names}
\end{table}
\begin{table}[t]
\centering
\caption{Comparison of the KP operators. In columns Stat the notation $X^+$ means the median of the measure is better than the one for variant $X$, $X^-$ means it is worse, and $X^*$ indicates no significant difference. Stat shows the results of Mann-Whitney U-test at significance level $5\%$}
\renewcommand{\tabcolsep}{2.5pt}
\renewcommand{\arraystretch}{0.9}
\begin{tabular}{l|cccc|cccc|cccc}
\toprule
             Int & DP &(1) & EA &(2) & DP &(1) & EA&(2) &DP&(1) &EA &(2)\\
\cmidrule(l{2pt}r{2pt}){2-3}
\cmidrule(l{2pt}r{2pt}){4-5}
\cmidrule(l{2pt}r{2pt}){6-7}
\cmidrule(l{2pt}r{2pt}){8-9}
\cmidrule(l{2pt}r{2pt}){10-11}
\cmidrule(l{2pt}r{2pt}){12-13}
            & $H$  & Stat &$H$ & Stat  & $H_e$ &Stat & $H_e$  &Stat & $H_i$& Stat& $H_i$& Stat        \\
\midrule
1&\hl{8.5}&$2^*$&8.3&$1^*$&5.4&$2^-$&\hl{5.7}&$1^+$&\hl{3}&$2^+$&2.6&$1^-$\\
2&\hl{9}&$2^+$&8.8&$1^-$&\hl{5.2}&$2^+$&5&$1^-$&3.8&$2^*$&3.8&$1^*$\\
3&\hl{9.5}&$2^+$&9.3&$1^-$&\hl{5.1}&$2^+$&5&$1^-$&4.3&$2^*$&\hl{4.4}&$1^*$\\
4&7.1&$2^*$&\hl{7.2}&$1^*$&\hl{5.3}&$2^+$&5.2&$1^-$&1.9&$2^*$&\hl{2}&$1^*$\\
5&\hl{8.7}&$2^+$&8.3&$1^-$&\hl{5.2}&$2^+$&5&$1^-$&\hl{3.4}&$2^+$&3.3&$1^-$\\
6&\hl{8.9}&$2^+$&8.8&$1^-$&\hl{5.2}&$2^+$&5&$1^-$&3.8&$2^*$&3.8&$1^*$\\
7&\hl{7.9}&$2^*$&7.8&$1^*$&5.3&$2^*$&5.3&$1^*$&2.5&$2^*$&2.5&$1^*$\\
8&8.6&$2^*$&8.6&$1^*$&5&$2^*$&5&$1^*$&3.5&$2^-$&\hl{3.6}&$1^+$\\
9&9.2&$2^*$&9.2&$1^*$&\hl{5.1}&$2^*$&5&$1^*$&4.1&$2^-$&4.1&$1^+$\\
10&9.8&$2^*$&9.8&$1^*$&\hl{6}&$2^+$&5.9&$1^-$&3.8&$2^-$&\hl{3.9}&$1^+$\\
11&10.7&$2^-$&10.7&$1^+$&6&$2^*$&6&$1^*$&4.7&$2^*$&4.7&$1^*$\\
12&8.6&$2^-$&\hl{8.8}&$1^+$&\hl{6}&$2^*$&5.9&$1^*$&2.7&$2^-$&\hl{2.8}&$1^+$\\
13&9.9&$2^*$&\hl{10}&$1^*$&\hl{6}&$2^+$&5.8&$1^-$&3.9&$2^-$&\hl{4.2}&$1^+$\\
14&9.4&$2^*$&9.4&$1^*$&5.9&$2^*$&5.9&$1^*$&3.4&$2^*$&\hl{3.5}&$1^*$\\
15&10.6&$2^*$&10.6&$1^*$&6&$2^*$&6&$1^*$&4.6&$2^-$&4.6&$1^+$\\
16&10.7&$2^*$&10.7&$1^*$&\hl{6.5}&$2^+$&6.4&$1^-$&4.3&$2^-$&4.4&$1^+$\\
17&10.1&$2^*$&10.1&$1^*$&\hl{6.5}&$2^+$&6.4&$1^-$&3.6&$2^*$&\hl{3.8}&$1^*$\\
18&10.7&$2^*$&10.7&$1^*$&\hl{6.5}&$2^+$&6.4&$1^-$&4.2&$2^-$&4.2&$1^+$\\
\bottomrule
\end{tabular}
\label{tbl:Res_OPR}
\end{table}
\section{Experimental Investigation}
\label{Sec:exp}
In this section, we conduct an comprehensive experimental investigation on the introduced framework to analyse the inter-dependency of the TTP's sub-problems in terms of structural diversity and find the best method to maximise it. First, we compare the two KP search operators, DP and $(1+1)$EA; then, we incorporate the $H$, $H_e$, and $H_i$ into the algorithm as the fitness function, and analyse the populations obtained. Finally, we conduct a comparison on the introduced framework with a recently introduced QD-based EA \cite{NikfarjamMap} in terms of structural diversity and robustness against small changes in availability of edges and items.  In terms of experimental setting, we used 18 TTP instances from \cite{PolyakovskiyB0MN14}, and the algorithms are terminated after $10000$ iterations. Table \ref{tab:names} shows the name of benchmarks instances. The internal termination criterion for the $(1+1)$EA is set to $2m$ based on preliminary experiments. We consider 10 independent runs for each algorithms on each test instances.    
\begin{table*}
\centering
\caption{Comparison of different fitness function (DP used as the KP operator). Stat shows the results of Kruskal-Wallis statistical test at significance level $5\%$ and Bonferroni correction. The notations are in line with Table \ref{tbl:Res_OPR}}
\renewcommand{\tabcolsep}{3pt}
\renewcommand{\arraystretch}{0.6}
\begin{tabular}{l|cccccc|cccccc|cccccc}
\toprule
               Ins & $H$ &(1) & $H_{e}$&(2) &$H_{i}$&(3) &$H$&(1) &$H_{e}$&(2) &$H_{i}$&(3) &$H$&(1) &$H_{e}$&(2) &$H_{i}$&(3) \\
\cmidrule(l{2pt}r{2pt}){2-3}
\cmidrule(l{2pt}r{2pt}){4-5}
\cmidrule(l{2pt}r{2pt}){6-7}
\cmidrule(l{2pt}r{2pt}){8-9}
\cmidrule(l{2pt}r{2pt}){10-11}
\cmidrule(l{2pt}r{2pt}){12-13}
\cmidrule(l{2pt}r{2pt}){14-15}
\cmidrule(l{2pt}r{2pt}){16-17}
\cmidrule(l{2pt}r{2pt}){18-19} 
            & $H$  & Stat &$H$ & Stat &$H$ & Stat & $H_e$ &Stat & $H_e$  &Stat& $H_e$ &Stat & $H_i$& Stat& $H_i$& Stat  & $H_i$& Stat      \\
\midrule
1&\hl{8.5}&$2^*3^+$&8.3&$1^*3^+$&7.5&$1^-2^-$&5.4&$2^-3^+$&\hl{5.8}&$1^+3^+$&4.8&$1^-2^-$&\hl{3}&$2^+3^+$&2.6&$1^-3^*$&2.7&$1^-2^*$\\
2&9&$2^*3^+$&\hl{9.1}&$1^*3^+$&8.6&$1^-2^-$&5.2&$2^-3^+$&\hl{5.4}&$1^+3^+$&4.8&$1^-2^-$&3.8&$2^+3^+$&3.7&$1^-3^*$&3.8&$1^-2^*$\\
3&9.5&$2^*3^+$&\hl{9.6}&$1^*3^+$&9.1&$1^-2^-$&5.1&$2^-3^+$&\hl{5.3}&$1^+3^+$&4.8&$1^-2^-$&4.3&$2^*3^*$&4.3&$1^*3^*$&4.3&$1^*2^*$\\
4&\hl{7.1}&$2^*3^+$&6.8&$1^*3^+$&6.4&$1^-2^-$&5.3&$2^*3^+$&\hl{5.4}&$1^*3^+$&4.8&$1^-2^-$&\hl{1.9}&$2^+3^*$&1.5&$1^-3^*$&1.6&$1^*2^*$\\
5&\hl{8.7}&$2^+3^+$&8.3&$1^-3^*$&7.7&$1^-2^*$&5.2&$2^-3^+$&\hl{5.4}&$1^+3^+$&4.8&$1^-2^-$&\hl{3.4}&$2^+3^+$&2.8&$1^-3^*$&2.9&$1^-2^*$\\
6&\hl{8.9}&$2^*3^+$&8.7&$1^*3^+$&8.3&$1^-2^-$&5.2&$2^-3^+$&\hl{5.3}&$1^+3^+$&4.8&$1^-2^-$&\hl{3.8}&$2^+3^+$&3.4&$1^-3^*$&3.5&$1^-2^*$\\
7&\hl{7.9}&$2^*3^+$&7.8&$1^*3^+$&7.3&$1^-2^-$&5.3&$2^*3^+$&\hl{5.4}&$1^*3^+$&4.8&$1^-2^-$&2.5&$2^+3^+$&2.4&$1^-3^*$&2.5&$1^-2^*$\\
8&8.6&$2^-3^+$&\hl{8.7}&$1^+3^+$&8.2&$1^-2^-$&5&$2^-3^+$&\hl{5.2}&$1^+3^+$&4.7&$1^-2^-$&3.5&$2^+3^+$&3.5&$1^-3^*$&3.5&$1^-2^*$\\
9&9.2&$2^*3^+$&\hl{9.3}&$1^*3^+$&8.8&$1^-2^-$&5.1&$2^*3^+$&\hl{5.2}&$1^*3^+$&4.7&$1^-2^-$&4.1&$2^+3^*$&4.1&$1^-3^*$&4.1&$1^*2^*$\\
10&9.8&$2^*3^+$&\hl{9.9}&$1^*3^+$&9.6&$1^-2^-$&6&$2^-3^+$&\hl{6.2}&$1^+3^+$&5.8&$1^-2^-$&3.8&$2^+3^*$&3.7&$1^-3^-$&3.8&$1^*2^+$\\
11&10.7&$2^*3^+$&\hl{10.8}&$1^*3^+$&10.5&$1^-2^-$&6&$2^*3^+$&\hl{6.1}&$1^*3^+$&5.8&$1^-2^-$&4.7&$2^+3^*$&4.7&$1^-3^*$&4.7&$1^*2^*$\\
12&8.6&$2^*3^+$&8.6&$1^*3^+$&8.5&$1^-2^-$&6&$2^*3^+$&6&$1^*3^+$&5.8&$1^-2^-$&2.7&$2^+3^*$&2.6&$1^-3^-$&2.7&$1^*2^+$\\
13&9.9&$2^*3^+$&9.9&$1^*3^+$&9.7&$1^-2^-$&6&$2^*3^+$&\hl{6.1}&$1^*3^+$&5.8&$1^-2^-$&3.9&$2^+3^*$&3.8&$1^-3^-$&3.9&$1^*2^+$\\
14&9.4&$2^*3^+$&9.4&$1^*3^+$&9.2&$1^-2^-$&5.9&$2^*3^+$&\hl{6}&$1^*3^+$&5.8&$1^-2^-$&3.4&$2^+3^*$&3.4&$1^-3^-$&3.4&$1^*2^+$\\
15&10.6&$2^*3^+$&10.6&$1^*3^+$&10.4&$1^-2^-$&6&$2^*3^+$&6&$1^*3^+$&5.8&$1^-2^-$&4.6&$2^*3^*$&4.6&$1^*3^*$&4.6&$1^*2^*$\\
16&10.7&$2^*3^+$&\hl{10.8}&$1^*3^+$&10.6&$1^-2^-$&6.5&$2^*3^+$&\hl{6.6}&$1^*3^+$&6.4&$1^-2^-$&\hl{4.3}&$2^+3^*$&4.2&$1^-3^*$&4.2&$1^*2^*$\\
17&\hl{10.1}&$2^*3^*$&9.9&$1^*3^*$&9.9&$1^*2^*$&6.5&$2^*3^+$&\hl{6.6}&$1^*3^+$&6.4&$1^-2^-$&3.6&$2^+3^*$&3.4&$1^-3^-$&3.6&$1^*2^+$\\
18&10.7&$2^*3^+$&\hl{10.8}&$1^*3^+$&10.6&$1^-2^-$&6.5&$2^*3^+$&\hl{6.6}&$1^*3^+$&6.4&$1^-2^-$&4.2&$2^+3^*$&4.2&$1^-3^-$&4.2&$1^*2^+$\\
\bottomrule
\end{tabular}
\label{tbl:Res_FitDp}
\end{table*}
\begin{table*}[t]
\centering
\caption{Comparison of different fitness function (EA used as the KP operator).The notations are in line with Table \ref{tbl:Res_FitDp}}
\renewcommand{\tabcolsep}{3pt}
\renewcommand{\arraystretch}{0.6}
\begin{tabular}{l|cccccc|cccccc|cccccc}
\toprule
               Ins & $H$ &(1) & $H_{e}$&(2) &$H_{i}$&(3) &$H$&(1) &$H_{e}$&(2) &$H_{i}$&(3) &$H$&(1) &$H_{e}$&(2) &$H_{i}$&(3) \\
\cmidrule(l{2pt}r{2pt}){2-3}
\cmidrule(l{2pt}r{2pt}){4-5}
\cmidrule(l{2pt}r{2pt}){6-7}
\cmidrule(l{2pt}r{2pt}){8-9}
\cmidrule(l{2pt}r{2pt}){10-11}
\cmidrule(l{2pt}r{2pt}){12-13}
\cmidrule(l{2pt}r{2pt}){14-15}
\cmidrule(l{2pt}r{2pt}){16-17}
\cmidrule(l{2pt}r{2pt}){18-19} 
            & $H$  & Stat &$H$ & Stat &$H$ & Stat & $H_e$ &Stat & $H_e$  &Stat& $H_e$ &Stat & $H_i$& Stat& $H_i$& Stat  & $H_i$& Stat      \\
\midrule
01&8.3&$2^*3^+$&8.3&$1^*3^+$&7.5&$1^-2^-$&5.7&$2^*3^+$&\hl{5.8}&$1^*3^+$&4.9&$1^-2^-$&2.6&$2^+3^*$&2.5&$1^-3^-$&2.6&$1^*2^+$\\
02&8.8&$2^*3^+$&\hl{9}&$1^*3^+$&8.4&$1^-2^-$&5&$2^-3^+$&\hl{5.3}&$1^+3^+$&4.7&$1^-2^-$&3.8&$2^+3^*$&3.7&$1^-3^-$&3.8&$1^*2^+$\\
03&9.3&$2^*3^+$&\hl{9.4}&$1^*3^+$&9&$1^-2^-$&5&$2^-3^+$&\hl{5.2}&$1^+3^+$&4.7&$1^-2^-$&\hl{4.4}&$2^+3^*$&4.2&$1^-3^-$&4.3&$1^*2^+$\\
04&\hl{7.2}&$2^+3^+$&6.9&$1^-3^+$&6.4&$1^-2^-$&5.2&$2^-3^+$&\hl{5.4}&$1^+3^+$&4.8&$1^-2^-$&\hl{2}&$2^+3^+$&1.5&$1^-3^*$&1.6&$1^-2^*$\\
05&\hl{8.3}&$2^*3^+$&8.1&$1^*3^+$&7.7&$1^-2^-$&5&$2^-3^*$&\hl{5.4}&$1^+3^+$&4.7&$1^*2^-$&\hl{3.3}&$2^+3^+$&2.7&$1^-3^*$&3&$1^-2^*$\\
06&\hl{8.8}&$2^+3^+$&8.6&$1^-3^*$&8.2&$1^-2^*$&5&$2^-3^+$&\hl{5.3}&$1^+3^+$&4.7&$1^-2^-$&\hl{3.8}&$2^+3^*$&3.3&$1^-3^-$&3.6&$1^*2^+$\\
07&7.8&$2^*3^+$&7.8&$1^*3^+$&7.3&$1^-2^-$&5.3&$2^*3^+$&\hl{5.4}&$1^*3^+$&4.8&$1^-2^-$&2.5&$2^+3^*$&2.4&$1^-3^*$&2.5&$1^*2^*$\\
08&8.6&$2^*3^+$&\hl{8.7}&$1^*3^+$&8.3&$1^-2^-$&5&$2^-3^+$&\hl{5.2}&$1^+3^+$&4.7&$1^-2^-$&3.6&$2^+3^*$&3.5&$1^-3^-$&3.6&$1^*2^+$\\
09&9.2&$2^*3^+$&\hl{9.3}&$1^*3^+$&8.8&$1^-2^-$&5&$2^-3^+$&\hl{5.2}&$1^+3^+$&4.7&$1^-2^-$&4.1&$2^+3^*$&4.1&$1^-3^-$&4.1&$1^*2^+$\\
10&9.8&$2^*3^+$&\hl{10}&$1^*3^+$&9.6&$1^-2^-$&5.9&$2^-3^+$&\hl{6.2}&$1^+3^+$&5.7&$1^-2^-$&3.9&$2^+3^*$&3.8&$1^-3^-$&3.9&$1^*2^+$\\
11&10.7&$2^*3^+$&\hl{10.8}&$1^*3^+$&10.5&$1^-2^-$&6&$2^*3^+$&\hl{6.1}&$1^*3^+$&5.7&$1^-2^-$&4.7&$2^*3^*$&4.7&$1^*3^-$&\hl{4.8}&$1^*2^+$\\
12&\hl{8.8}&$2^*3^*$&8.7&$1^*3^*$&8.7&$1^*2^*$&5.9&$2^*3^+$&\hl{6}&$1^*3^+$&5.7&$1^-2^-$&2.8&$2^+3^*$&2.6&$1^-3^-$&\hl{2.9}&$1^*2^+$\\
13&\hl{10}&$2^+3^*$&9.9&$1^-3^*$&9.9&$1^*2^*$&5.8&$2^-3^+$&\hl{6.1}&$1^+3^+$&5.7&$1^-2^-$&4.2&$2^+3^*$&3.8&$1^-3^-$&4.2&$1^*2^+$\\
14&9.4&$2^*3^+$&9.4&$1^*3^+$&9.2&$1^-2^-$&5.9&$2^-3^+$&\hl{6}&$1^+3^+$&5.8&$1^-2^-$&3.5&$2^+3^*$&3.4&$1^-3^-$&3.5&$1^*2^+$\\
15&10.6&$2^*3^+$&10.6&$1^*3^+$&10.3&$1^-2^-$&6&$2^*3^+$&6&$1^*3^+$&5.7&$1^-2^-$&4.6&$2^+3^*$&4.6&$1^-3^*$&4.6&$1^*2^*$\\
16&10.7&$2^*3^*$&\hl{10.8}&$1^*3^+$&10.7&$1^*2^-$&6.4&$2^-3^+$&\hl{6.6}&$1^+3^+$&6.3&$1^-2^-$&4.4&$2^+3^*$&4.2&$1^-3^-$&4.4&$1^*2^+$\\
17&10.1&$2^+3^*$&9.9&$1^-3^-$&10.1&$1^*2^+$&6.4&$2^-3^*$&\hl{6.6}&$1^+3^+$&6.3&$1^*2^-$&\hl{3.8}&$2^+3^*$&3.3&$1^-3^-$&3.7&$1^*2^+$\\
18&10.7&$2^*3^*$&10.7&$1^*3^+$&10.6&$1^*2^-$&6.4&$2^*3^+$&\hl{6.6}&$1^*3^+$&6.3&$1^-2^-$&4.2&$2^+3^*$&4.2&$1^-3^-$&4.2&$1^*2^+$\\
\bottomrule
\end{tabular}
\label{tbl:Res_FitOne}
\end{table*}
\subsection{Comparison in KP search operators operators}

In this section,  we compute two set of solutions for each test instance, one by use of DP and another with $(1+1)$EA, and scrutinise the diversity of the sets. Here, $H$ serves as fitness function and $\alpha$ and $\mu$ are set to $0.1$ and $50$, respectively. Table \ref{tbl:Res_OPR} summarises the results. As Table \ref{tbl:Res_OPR} shows, the use of DP results in a population with higher diversity in edges ($H_e$), while $H_i$ is higher in the population obtained from $(1+1)$EA in most of cases. Turning to overall diversity ($H$), the use of $DP$ brings about populations with higher diversity in 4 out of 18 cases. On the other hand, there are 2 cases, that ($1+1$)EA outperforms the DP. There are found no significant differences in overall diversity for the rest of the test instances. One may ask the question why using DP results in a higher $H_e$, while EAX is used to generate new tours in both competitors. One explanation is that DP compute the same packing list for two identical tours. This is while, $(1+1)$EA can generate different packing lists which results in a higher $H_i$ but a lower $H_e$.

\subsection{Comparison in fitness functions}
\begin{figure*}
\centering

\includegraphics[width=.30\columnwidth]{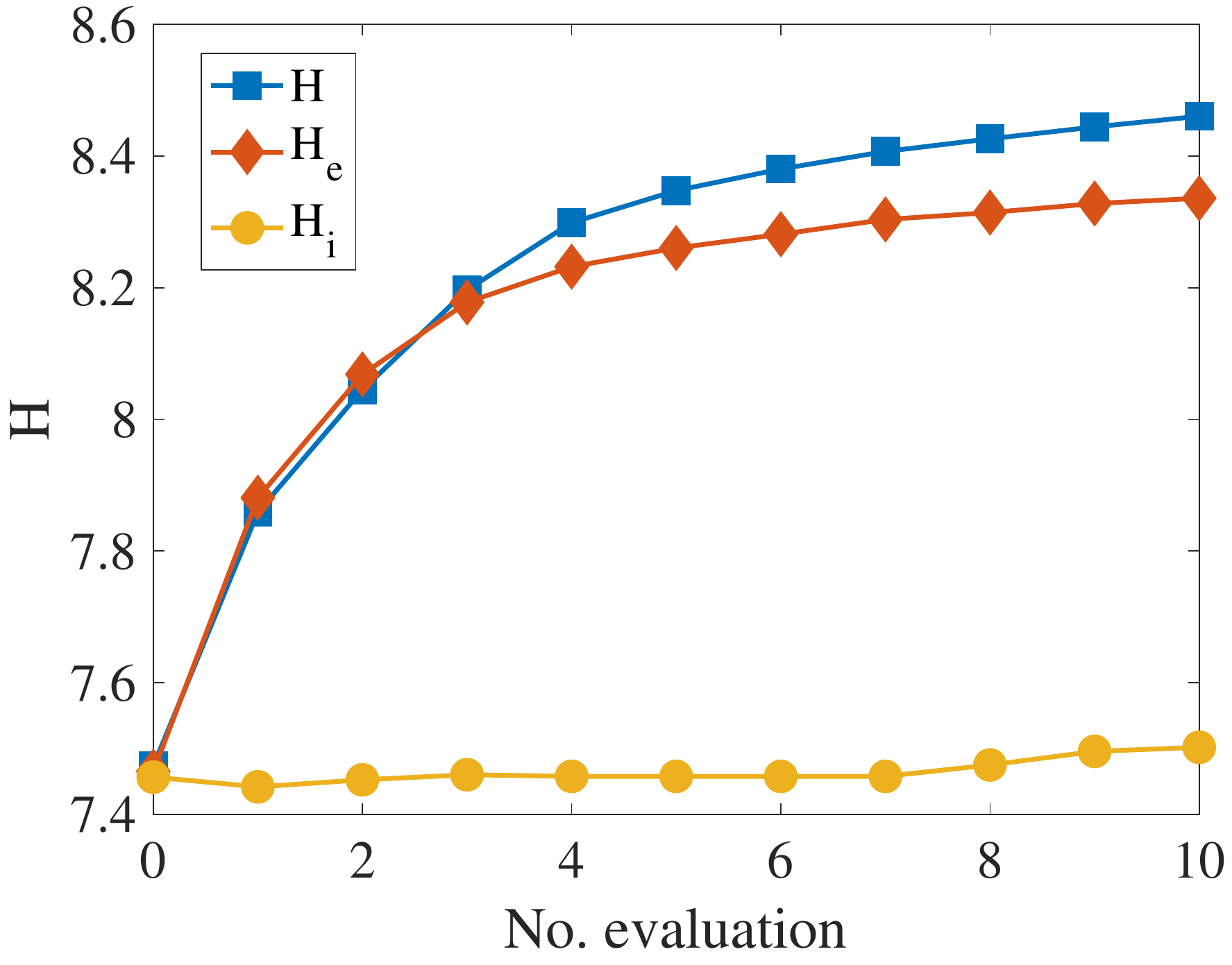}
\hskip5pt
\includegraphics[width=.30\columnwidth]{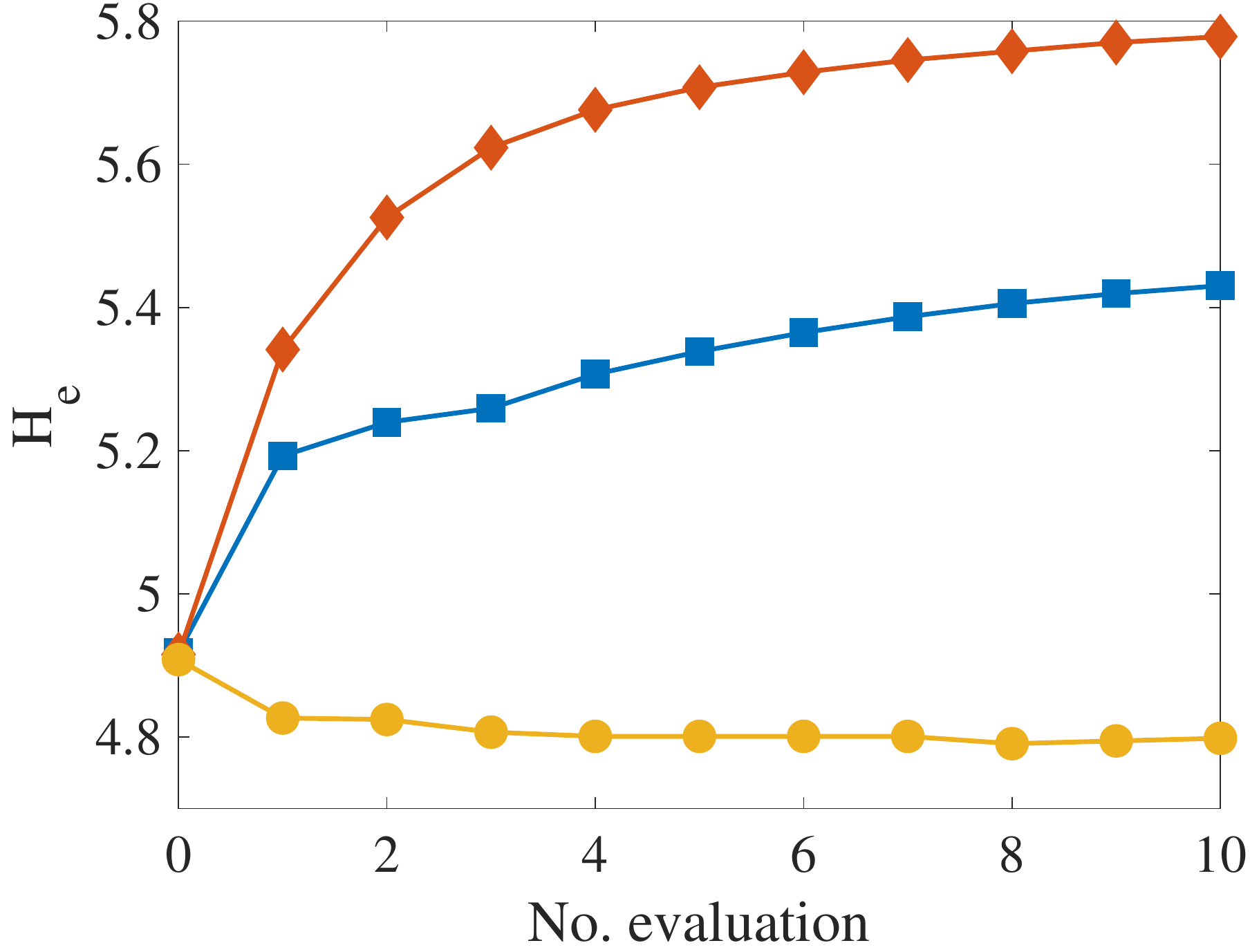}
\hskip5pt
\includegraphics[width=.30\columnwidth]{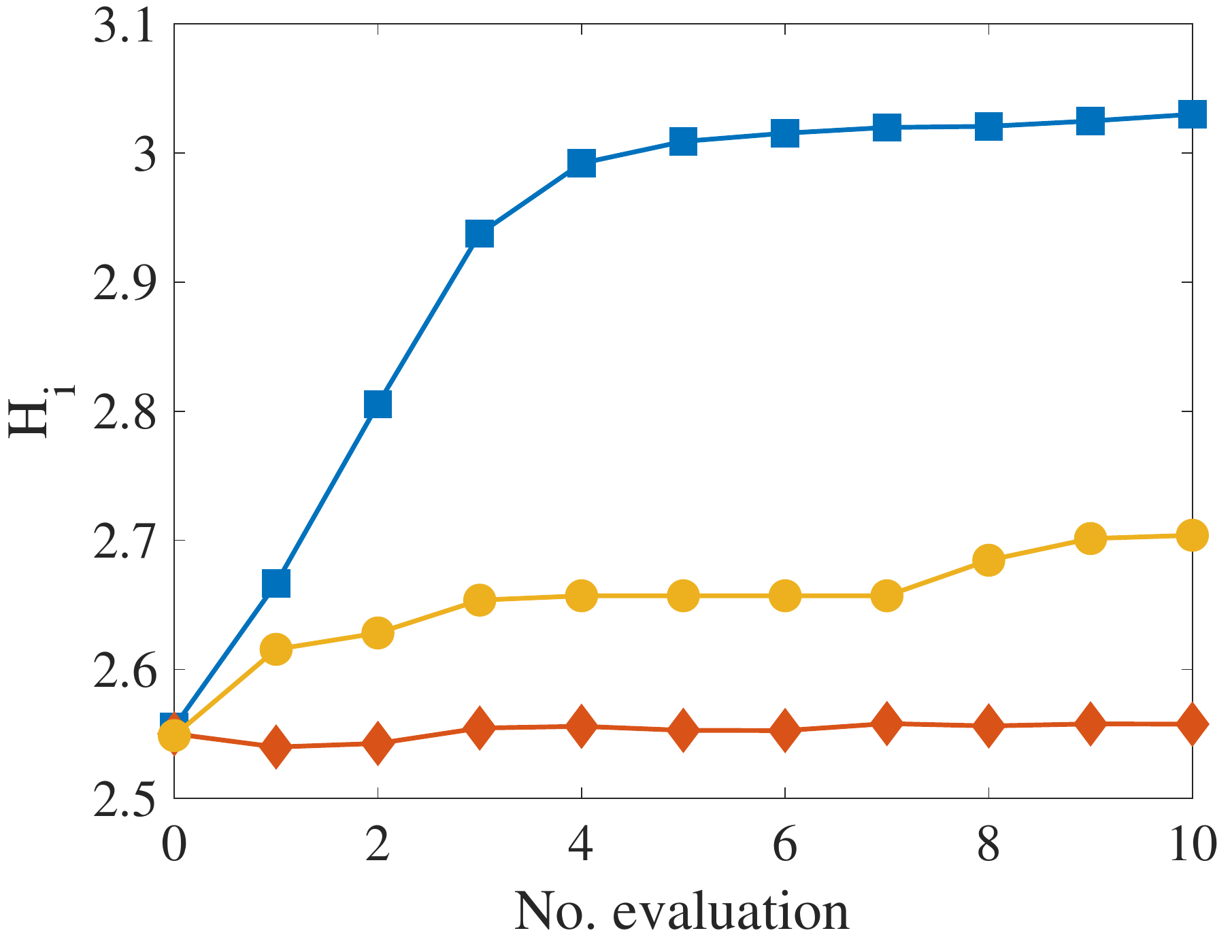}
\hskip5pt
\includegraphics[width=.30\columnwidth]{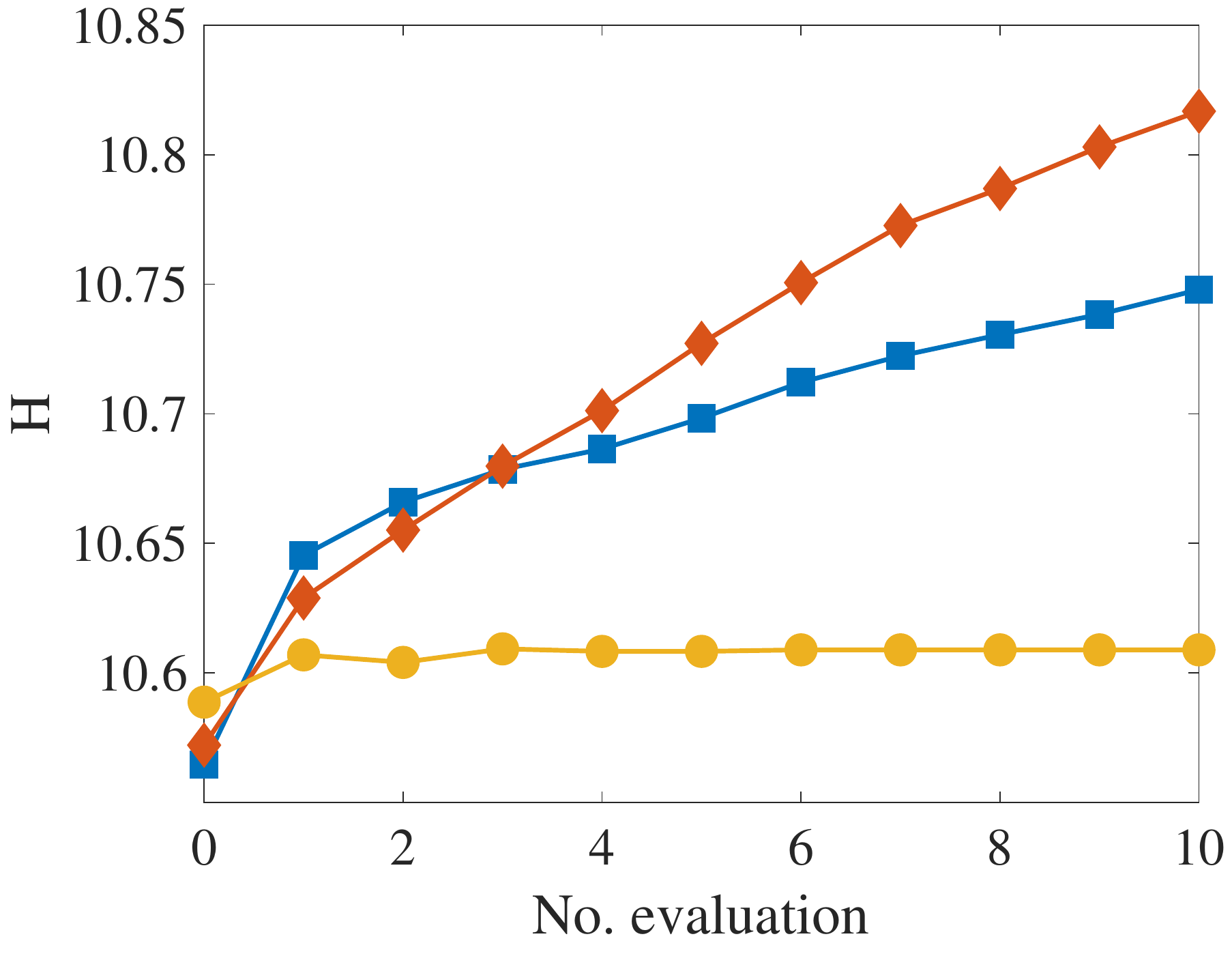}
\hskip5pt
\includegraphics[width=.30\columnwidth]{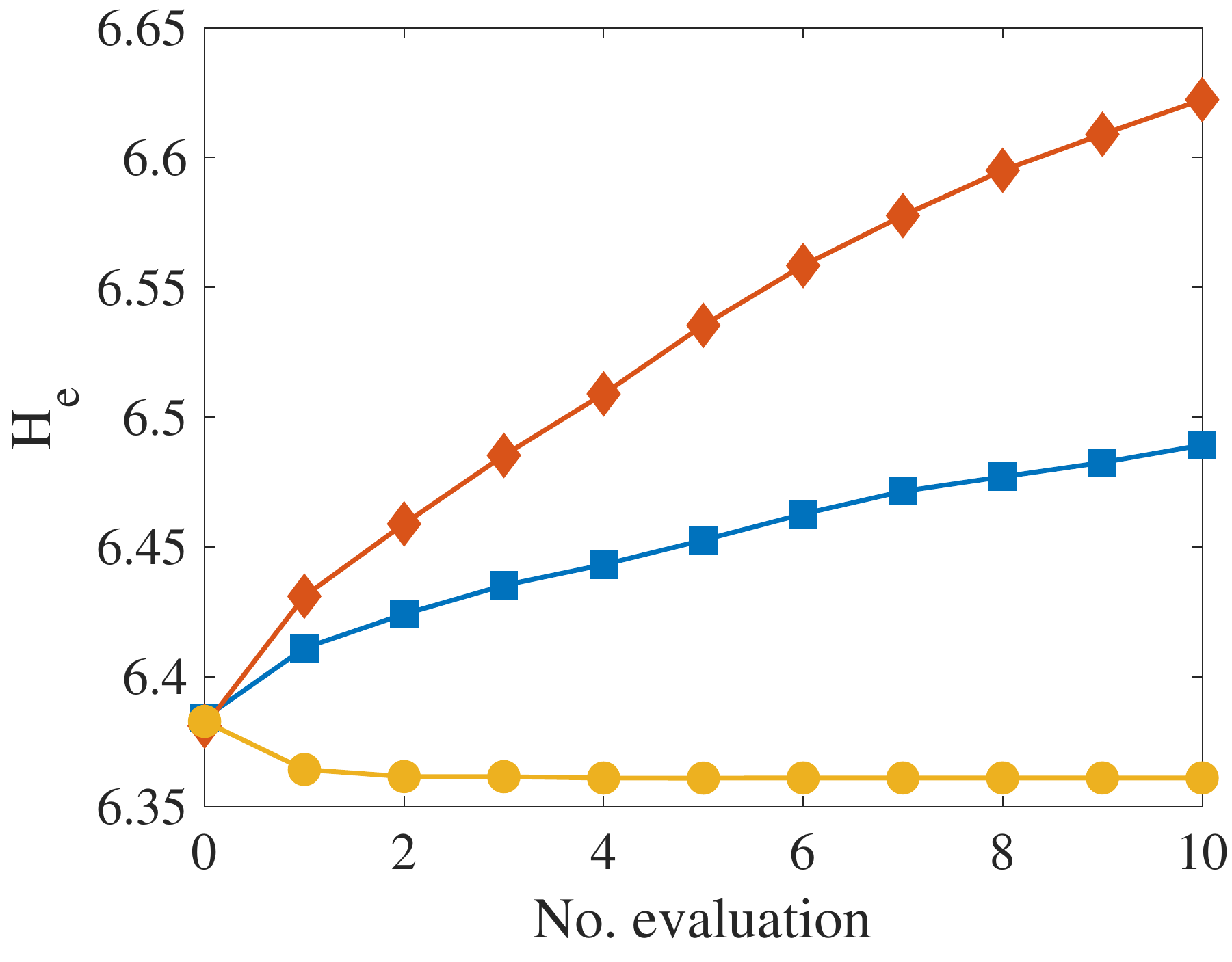}
\hskip5pt
\includegraphics[width=.30\columnwidth]{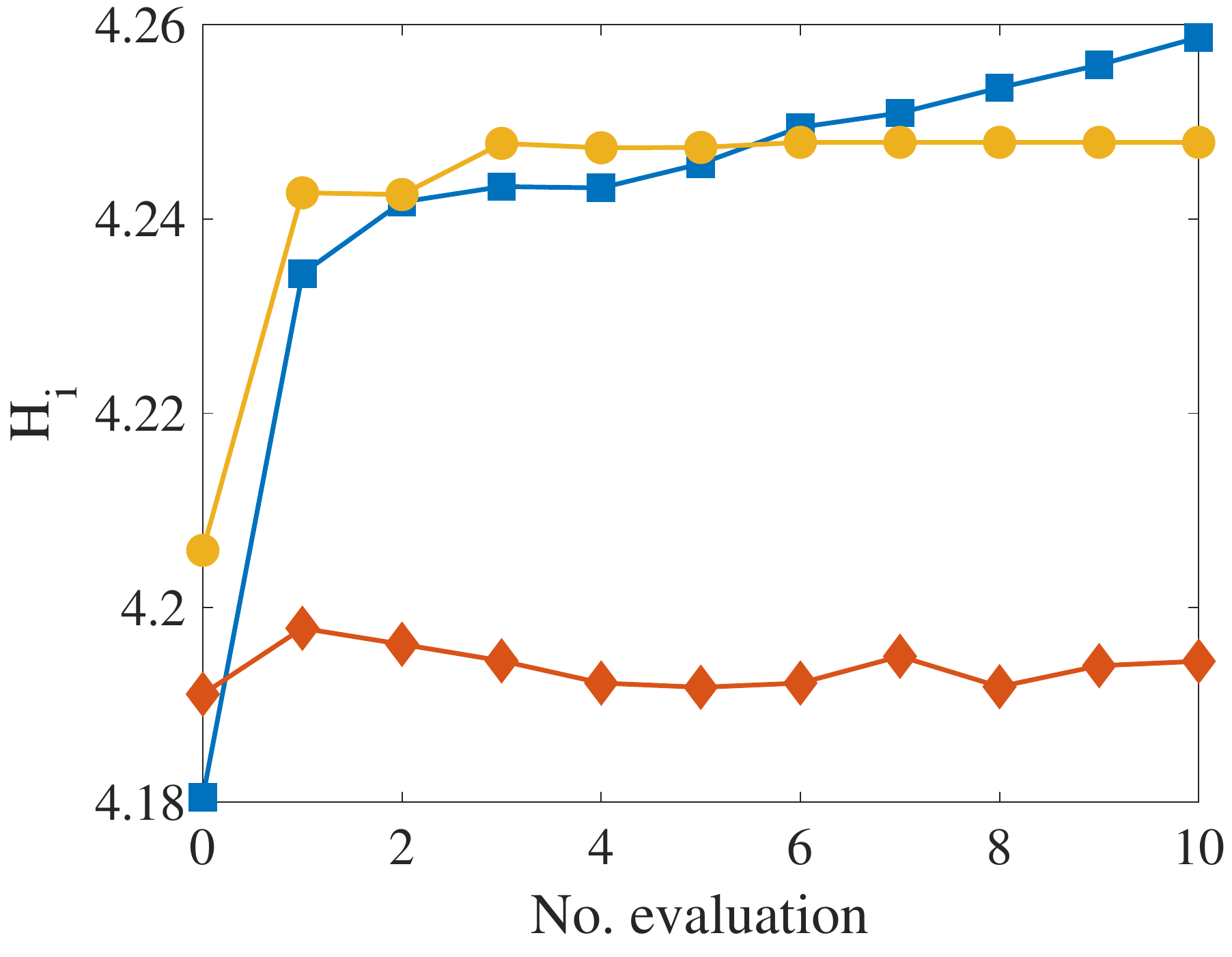}

\caption{Representation of trajectories of incorporation of the different fitness functions, $H$, $H_e$, and $H_i$ over test instance 1 (first row), and test instance 16 (second row)}
\label{fig:Trd}
\end{figure*}
Next, we investigate the use of $H_e$ or $H_i$ as the fitness functions instead of $H$. Table \ref{tbl:Res_FitDp} compares three algorithms using the fitness functions where DP is used as KP operator. The table shows there are no significant differences in overall diversity when either of $H$ or $H_{e}$ serve as the fitness function. However, the EA using item diversity ($H_i$) results in the populations with an overall entropy significantly less than the other EAs. It also gets outperformed by the EA using $H$ in terms of items diversity. This is because the introduced framework is a bi-level optimisation procedure where it generates a tour first; then, it computes the packing list based on the tour. Therefore, the use of diversity in edge can aid in increasing the diversity of items, especially where the EA uses DP. Figure \ref{fig:Trd} depicts the trajectories of the EAs using the three fitness functions over $10000$ iterations in the test instances 1 and 16. The figure explicitly confirms the previous observations; using $H_i$ as the fitness function makes the EA incapable of maximising overall and edge diversity. It also gets outperformed in terms of item entropy $H_i$. On the other hand, incorporating $H_e$ as the fitness functions results in decent overall and edge diversity. However, it can not increase the item entropy. Figure \ref{fig:Trd} also shows the EAs using $H$ and $H_e$ as fitness function do not converge in $10000$ iterations for the test instance 16 (a280\_n279\_bounded-strongly-corr\_01). Overall, if we aim to increase the total or edge diversity, using $H_e$ as the fitness function would be better. This is because we achieve similar total diversity with the use $H_e$, but it results in higher entropy in the edges, and more importantly, it requires less calculation. However, $H$ works the best if we focus on the diversity of items or a more balanced diversity between items and edges.          

Now, we conduct the same experiments with $(1+1)$EA to observe the changes in the results. Table \ref{tbl:Res_FitOne} summarises the results for this round of experiments. Here, one can observe that using $H$ slightly outperforms $H_e$ if we aim for total diversity. The underlying reason is that DP is an exact algorithm that results in the same packing list for identical tours. Thus, identical tours have no contribution to the diversity of edges or items. This is while the $(1+1)$EA can return different packing lists for identical tours and contribute to the diversity of items and overall diversity. Thus, overall diversity is slightly higher when $H$ is used as the fitness function when we incorporate $(1+1)$EA as the inner algorithm.      

\begin{table}[t]
\centering
\caption{Comparison of the EDO and QD (DP used as the KP operator). The notations are in line with Table \ref{tbl:Res_OPR}.}
\renewcommand{\tabcolsep}{1.5pt}
\renewcommand{\arraystretch}{0.9}
\begin{tabular}{l|cccc|cccc|cccc}
\toprule
             Int & EDO &(1) & QD &(2) & EDO &(1) & QD&(2) & EDO &(1) & QD &(2)\\
\cmidrule(l{2pt}r{2pt}){2-3}
\cmidrule(l{2pt}r{2pt}){4-5}
\cmidrule(l{2pt}r{2pt}){6-7}
\cmidrule(l{2pt}r{2pt}){8-9}
\cmidrule(l{2pt}r{2pt}){10-11}
\cmidrule(l{2pt}r{2pt}){12-13}
            & $H$  & Stat & $H$ & Stat  & $H_e$ &Stat & $H_e$  &Stat & $H_i$& Stat& $H_i$& Stat        \\
\midrule
01&\hl{9.1}&$2^+$&8.1&$1^-$&\hl{5.9}&$2^+$&5.1&$1^-$&\hl{3.2}&$2^+$&3&$1^-$\\
02&\hl{9.8}&$2^+$&9.1&$1^-$&\hl{5.6}&$2^+$&5.1&$1^-$&\hl{4.2}&$2^+$&3.9&$1^-$\\
03&\hl{10.2}&$2^+$&9.5&$1^-$&\hl{5.5}&$2^+$&5.1&$1^-$&\hl{4.6}&$2^+$&4.4&$1^-$\\
04&\hl{9.1}&$2^+$&7.7&$1^-$&\hl{5.9}&$2^+$&5.2&$1^-$&\hl{3.2}&$2^+$&2.5&$1^-$\\
05&\hl{9.7}&$2^+$&8.7&$1^-$&\hl{5.7}&$2^+$&5.1&$1^-$&\hl{4}&$2^+$&3.5&$1^-$\\
06&\hl{10.3}&$2^+$&9.2&$1^-$&\hl{5.8}&$2^+$&5.1&$1^-$&\hl{4.5}&$2^+$&4.1&$1^-$\\
07&\hl{8.6}&$2^+$&7.8&$1^-$&\hl{5.8}&$2^+$&5.1&$1^-$&\hl{2.8}&$2^+$&2.7&$1^-$\\
08&\hl{9.3}&$2^+$&8.8&$1^-$&\hl{5.5}&$2^+$&5.1&$1^-$&\hl{3.8}&$2^+$&3.7&$1^-$\\
09&\hl{9.9}&$2^+$&9.3&$1^-$&\hl{5.6}&$2^+$&5.1&$1^-$&\hl{4.3}&$2^+$&4.2&$1^-$\\
10&\hl{10}&$2^+$&9.9&$1^-$&\hl{6.1}&$2^+$&6&$1^-$&3.9&$2^*$&3.9&$1^*$\\
11&\hl{11.1}&$2^+$&11&$1^-$&\hl{6.2}&$2^+$&6&$1^-$&5&$2^*$&5&$1^*$\\
12&\hl{9.6}&$2^+$&9.5&$1^-$&\hl{6.1}&$2^+$&6&$1^-$&3.5&$2^*$&3.5&$1^*$\\
13&\hl{10.7}&$2^*$&10.6&$1^*$&\hl{6.1}&$2^+$&6&$1^-$&4.5&$2^*$&\hl{4.6}&$1^*$\\
14&\hl{9.8}&$2^+$&9.5&$1^-$&\hl{6.3}&$2^+$&6&$1^-$&3.5&$2^-$&\hl{3.6}&$1^+$\\
15&\hl{10.9}&$2^+$&10.8&$1^-$&\hl{6.3}&$2^+$&6&$1^-$&4.7&$2^-$&\hl{4.8}&$1^+$\\
16&10.9&$2^-$&\hl{11.2}&$1^+$&6.5&$2^-$&\hl{6.6}&$1^+$&4.4&$2^*$&\hl{4.5}&$1^*$\\
17&\hl{10.8}&$2^*$&10.7&$1^*$&6.5&$2^-$&\hl{6.7}&$1^+$&\hl{4.3}&$2^+$&4&$1^-$\\
18&10.8&$2^-$&\hl{11.1}&$1^+$&6.5&$2^-$&\hl{6.7}&$1^+$&4.3&$2^-$&\hl{4.4}&$1^+$\\
\end{tabular}
\label{tbl:Res_MapDP}
\end{table}
\begin{table}[t]
\centering
\caption{Comparison of the EDO and QD (EA used as the KP operator). The notations are in line with Table \ref{tbl:Res_OPR}.}
\renewcommand{\tabcolsep}{1.5pt}
\renewcommand{\arraystretch}{0.9}
\begin{tabular}{l|cccc|cccc|cccc}
\toprule
             Int & EDO &(1) & QD &(2) & EDO &(1) & QD&(2) & EDO &(1) & QD &(2)\\
\cmidrule(l{2pt}r{2pt}){2-3}
\cmidrule(l{2pt}r{2pt}){4-5}
\cmidrule(l{2pt}r{2pt}){6-7}
\cmidrule(l{2pt}r{2pt}){8-9}
\cmidrule(l{2pt}r{2pt}){10-11}
\cmidrule(l{2pt}r{2pt}){12-13}
            & $H$  & Stat & $H$ & Stat  & $H_e$ &Stat & $H_e$  &Stat & $H_i$& Stat& $H_i$& Stat        \\
\midrule
01&\hl{9}&$2^+$&8.1&$1^-$&\hl{5.8}&$2^+$&5.1&$1^-$&\hl{3.2}&$2^+$&3&$1^-$\\
02&\hl{9.7}&$2^+$&9.1&$1^-$&\hl{5.5}&$2^+$&5.1&$1^-$&\hl{4.2}&$2^+$&3.9&$1^-$\\
03&\hl{10.1}&$2^+$&9.5&$1^-$&\hl{5.5}&$2^+$&5.1&$1^-$&\hl{4.6}&$2^+$&4.4&$1^-$\\
04&\hl{9}&$2^+$&7.7&$1^-$&\hl{5.8}&$2^+$&5.2&$1^-$&\hl{3.2}&$2^+$&2.5&$1^-$\\
05&\hl{9.6}&$2^+$&8.7&$1^-$&\hl{5.5}&$2^+$&5.1&$1^-$&\hl{4.1}&$2^+$&3.5&$1^-$\\
06&\hl{10.2}&$2^+$&9.2&$1^-$&\hl{5.6}&$2^+$&5.1&$1^-$&\hl{4.6}&$2^+$&4.1&$1^-$\\
07&\hl{8.5}&$2^+$&7.8&$1^-$&\hl{5.8}&$2^+$&5.1&$1^-$&\hl{2.8}&$2^+$&2.7&$1^-$\\
08&\hl{9.2}&$2^+$&8.8&$1^-$&\hl{5.4}&$2^+$&5.1&$1^-$&\hl{3.8}&$2^+$&3.7&$1^-$\\
09&\hl{9.9}&$2^+$&9.3&$1^-$&\hl{5.5}&$2^+$&5.1&$1^-$&\hl{4.3}&$2^+$&4.2&$1^-$\\
10&\hl{10}&$2^+$&9.9&$1^-$&6&$2^+$&6&$1^-$&\hl{4}&$2^+$&3.9&$1^-$\\
11&\hl{11.1}&$2^*$&11&$1^*$&6&$2^*$&6&$1^*$&5&$2^*$&5&$1^*$\\
12&\hl{9.6}&$2^*$&9.5&$1^*$&5.9&$2^-$&\hl{6}&$1^+$&\hl{3.7}&$2^+$&3.5&$1^-$\\
13&10.6&$2^*$&10.6&$1^*$&5.9&$2^-$&\hl{6}&$1^+$&\hl{4.7}&$2^+$&4.6&$1^-$\\
14&\hl{9.7}&$2^+$&9.5&$1^-$&\hl{6.1}&$2^+$&6&$1^-$&3.5&$2^*$&\hl{3.6}&$1^*$\\
15&\hl{10.9}&$2^+$&10.8&$1^-$&\hl{6.3}&$2^+$&6&$1^-$&4.6&$2^-$&\hl{4.8}&$1^+$\\
16&11&$2^-$&\hl{11.2}&$1^+$&6.4&$2^-$&\hl{6.6}&$1^+$&\hl{4.6}&$2^*$&4.5&$1^*$\\
17&\hl{10.9}&$2^+$&10.7&$1^-$&6.4&$2^-$&\hl{6.7}&$1^+$&\hl{4.5}&$2^+$&4&$1^-$\\
18&10.8&$2^-$&\hl{11.1}&$1^+$&6.4&$2^-$&\hl{6.7}&$1^+$&4.4&$2^-$&4.4&$1^+$\\
\end{tabular}
\label{tbl:Res_mapOne}
\end{table}

\begin{table}[t]
\centering
\caption{Comparison of the robustness of the populations obtained from the EDO-based EA (1) and the QD-based EA (2).  The $E$ and $I$ denotes the percentage of times the population has at least one alternative for the eliminated edges and item, respectively. The Stat notations are in line with Table \ref{tbl:Res_OPR}.}
\renewcommand{\tabcolsep}{4pt}
\renewcommand{\arraystretch}{0.9}
\begin{tabular}{l|cccc|cccc}
\toprule
             Int & EDO &(1) & QD &(2) & EDO &(1) & QD&(2) \\
\cmidrule(l{2pt}r{2pt}){2-3}
\cmidrule(l{2pt}r{2pt}){4-5}
\cmidrule(l{2pt}r{2pt}){6-7}
\cmidrule(l{2pt}r{2pt}){8-9}
            & E  & Stat &E & Stat      & I &Stat &I  &Stat        \\
              
\midrule
1&\hl{99.6}&$2^+$&87.5&$1^-$&\hl{70}&$2^+$&51.6&$1^-$\\
2&\hl{98.2}&$2^+$&92&$1^-$&\hl{60.7}&$2^+$&43.9&$1^-$\\
3&\hl{97.3}&$2^+$&85.5&$1^-$&\hl{56.2}&$2^+$&43.6&$1^-$\\
4&\hl{99.8}&$2^+$&90.2&$1^-$&\hl{51.2}&$2^+$&36.8&$1^-$\\
5&\hl{99.8}&$2^+$&89&$1^-$&\hl{41.6}&$2^+$&32.5&$1^-$\\
6&\hl{99.6}&$2^+$&86.9&$1^-$&\hl{43.8}&$2^+$&33.2&$1^-$\\
7&\hl{98.4}&$2^+$&90.2&$1^-$&\hl{30.4}&$2^+$&26.4&$1^-$\\
8&\hl{98.2}&$2^+$&85.1&$1^-$&\hl{28.7}&$2^+$&22.5&$1^-$\\
9&\hl{99}&$2^+$&89.8&$1^-$&\hl{28.6}&$2^+$&26.1&$1^-$\\
10&\hl{64.1}&$2^+$&54.4&$1^-$&27.7&$2^-$&\hl{31.1}&$1^+$\\
11&\hl{61.5}&$2^+$&49.9&$1^-$&30.8&$2^*$&\hl{34.9}&$1^*$\\
12&\hl{67}&$2^+$&54.6&$1^-$&25.4&$2^-$&\hl{28.9}&$1^+$\\
13&\hl{68.8}&$2^+$&54&$1^-$&23.5&$2^-$&\hl{26.2}&$1^+$\\
14&\hl{67.8}&$2^+$&47&$1^-$&7.7&$2^-$&\hl{15.5}&$1^+$\\
15&\hl{71.4}&$2^+$&52.8&$1^-$&7.7&$2^-$&\hl{18.4}&$1^+$\\
16&29&$2^-$&\hl{67.4}&$1^+$&24.8&$2^-$&\hl{35.8}&$1^+$\\
17&31&$2^-$&\hl{74.4}&$1^+$&\hl{31}&$2^+$&26.5&$1^-$\\
18&33.4&$2^-$&\hl{76.3}&$1^+$&16.7&$2^-$&\hl{22}&$1^+$\\
\end{tabular}
\label{tbl:Res_Rob}
\end{table}
\subsection{Comparison of EDO and QD}
We compare the introduced EDO-based framework with the QD-based EA in this section. We first run the QD-based EA for $10000$ iterations. Then, we set the quality threshold to the minimum quality found in the population obtained by the EA and set $\mu$ to the size of the set of solutions obtained. Having set the input parameters, we run the introduced EDO-based algorithm for the same number of iterations. Finally, we compare the two populations in terms of structural diversity ($H$, $H_e$, and $H_i$). 

In line with the previous section, we first employ DP as the KP operator; then, $(1+1)$EA is replaced with DP to analyse the impact of using different KP search operators in the results. Table \ref{tbl:Res_MapDP} shows the results when DP is employed. Compared to the QD-based algorithm, the introduced EA results in a higher $H$, $H_e$, and $H_i$ in 14, 15, and 9 cases out of 18, respectively. Table \ref{tbl:Res_mapOne} summarises the results when $(1+1)$EA is used as the KP operator. The results are almost in line with Table \ref{tbl:Res_MapDP}. Here, the performance of EDO-based EA improves in increasing entropy of items, while it deteriorates in overall and edge diversity. The table shows that the number of cases in favour of EDO-based EA increases to 13 cases taking $H_i$ into account. On the other hand, there is a fall of 1 and 3 cases in terms of $H$ and $H_i$, respectively. 

Furthermore, we conduct an experiment to test the robustness of populations obtained from the EDO and QD-based EAs against changes in the availability of edges and items. In this series of experiments, we make an edge of the best solution of the population unavailable and look into the population to check if there is a solution not using the excluded edge. For items, we look for solutions behaving the opposite of the best solution. For example, if item $i$ is included in the packing list of the best solution, we check if there is a solution excluding the item $i$, and vice versa. We repeat the experiments for all edges and items of the best solution. Table \ref{tbl:Res_Rob} summarises the results of the robustness experiment. The results show the EDO-based EA results in more robust sets of solutions. In the edges entropy $H_e$, it strongly outperforms the QD-based EA in 15 out of 18 test instances, while the figure is 10 for the entropy of items $H_i$. The results of the small instances (the first 9) where the EDO-based EA converges in $10000$ iterations indicate that EDO-based EA can provide a highly robust set of solutions if given sufficient time. Also, we can improve the robustness in edges if we alter the focus on the diversity of edges by using $H_e$ as the fitness function; however, the robustness in items is likely to decrease in this case.      
\section{Conclusion}
We introduced a framework to generate a set of high-quality TTP solutions differing in structural diversity. We examined the inter-dependency of TTP's sub-problems, TSP and KP, and determined the best method to achieve a highly diverse set of solutions. Moreover, We empirically analysed the introduced framework and compared the results with a recently-developed QD-based algorithm in terms of diversity. The results showed a considerable improvement in the diversity of the population compared to the QD-based algorithm. Finally, we conduct a simulation test to evaluate the robustness of the population obtained from the two frameworks.

For future study, it is intriguing to incorporate indicators from multi-objective optimisation frameworks into the algorithm to focus on diversities of edges and items and compare them to the incumbent method. Moreover, several multi-component real-world problems such as patient admission scheduling problem and vehicle rooting problem, can be found in the literature, where a set of diverse solutions is beneficial.   

\section{Acknowledgements}
This work has been supported by the Australian Research Council (ARC) through grants DP190103894, FT200100536, and by the South Australian Government through the Research Consortium “Unlocking Complex Resources through Lean Processing”.
\bibliographystyle{abbrvnat}
\bibliography{main}
\pagebreak

\end{document}